\documentclass[11pt]{article}
\usepackage{eamt26}
\usepackage{float}
\usepackage{times}
\usepackage{url}
\usepackage{latexsym}
\copyrightfn{© 2026 The authors. This article is licensed under a Creative Commons 4.0 licence, no derivative works, attribution, CC-BY-ND.}
\usepackage{graphicx}
\usepackage[small,bf]{caption} 
\title{On the Use of LLMs for Specialised Terminology:\\A Good Alternative to Corpora?}

\author{
  Joachim Minder\textsuperscript{1, 2} \and 
  Guillaume Wisniewski\textsuperscript{2} \and
  Natalie Kübler\textsuperscript{1} \\
  \textsuperscript{1}Université Paris Cité, ALTAE, F-75013 Paris, France \\
  \textsuperscript{2}Université Paris Cité, CNRS, Laboratoire de linguistique formelle, F-75013 Paris, France
}

\date{}

\begin{document}
\maketitle
\begin{abstract}
\fontsize{10pt}{11pt}\selectfont
  Specialised translation relies on the use of documentary and terminological resources, including corpora. These resources are particularly useful for terminology. However, their compilation and exploitation have several limitations: they require time, technical skills and access to data that can be difficult to collect. This study examines the extent to which LLMs can assist specialised translators in finding equivalents from English to French. We evaluate four proprietary models, GPT-4o, GPT-5.2, Claude Sonnet 4.5 and DeepSeek, in two specialised domains, Earth, Environmental and Planetary Sciences (EEPS) and Natural Language Processing (NLP). The experiment is based on 80 terms per domain and compares two prompting strategies: a terminology and a translation mode. The results highlight clear differences between models, prompting strategies and, to a lesser extent, domains. Claude Sonnet 4.5 achieves the best results in the most favourable configuration, while DeepSeek stands out for its greater stability.  Analysis of confidence estimates also shows that they are only a partial indicator of terminological accuracy. Overall, the findings suggest that LLMs can be useful tools for specialised translators, but cannot, at this stage, replace specialised corpora. This research therefore paves the way for future work on the real practical usefulness of LLMs for specialised translators in work and educational contexts.
\end{abstract}

\section{Introduction}

Specialised translation refers to the translation of texts that are specific to a particular field of knowledge, such as Earth sciences, computer science, law, etc. This type of translation, also known as pragmatic translation, has significant economic implications, as it is used in high-stakes industries.

The main challenges associated with specialised translation relate to aspects of Language for Specific Purposes (LSP). LSP is a language common to a group of specialists that serves the interests of that group \cite{bowker-pearson,basturkmen-elder,kies-hall-moore,gledhill-kubler}. Specialised translation--regardless of the domain--presents significant challenges, mainly in terms of terminology \cite{garcia-leon-arauz}. Each domain is characterised by its own conceptual network and, consequently, its own terminology \cite{scarpa-specialised}. For example, the terms \textit{fault creep}, \textit{effusive volcano} and \textit{very low-frequency earthquake} are part of the terminology of volcanology (and more broadly, Earth sciences). On the other hand, \textit{neural network}, \textit{large language model} and \textit{text mining} refer to natural language processing and computer science. However, some terms may have a different meaning in different specialised languages, such as \textit{cloud}, which does not refer to the same concept in Earth sciences as it does in computer science. 

Consequently, specialised translation, even just in terms of bilingual terminology research, is a complex, cognitively demanding and time-consuming task for translators, whether they are still learners or already experienced professionals:
“Pragmatically translating LSPs requires not only knowledge of the source and target cultures in general, but also knowledge of very specific areas. Even a very well-educated translator may not know the terminology, phraseology, or even grammar of a particular [specialised] domain” \cite{kubler:hal-01134954}

When it comes to terminology (but also other aspects such as phraseology, collocations, discursive conventions, etc.), there are several tools available to translators and translation learners. The simplest tools are arguably term bases and bilingual glossaries. However, these are far from exhaustive and never cover all the terms in a given field. To complement this method, specialised translators must turn to corpora. Studies on the use of corpora for translation and translation teaching began in the late 1990s, notably with Baker \shortcite{baker-corpora} and Aston \shortcite{aston-corpus}. Specialised translation is particularly affected by the need to use corpora \cite{kubler:hal-01134954,kubler-et-al-corpus-use,bernardini2022corpora,granger_lefer2022cbts}. Corpora are useful for acquiring knowledge in these specialised domains, finding linguistic information, and searching for terminological equivalents \cite{kubler-et-al-corpus-use}. Two main types of corpora can be compiled to assist specialised translation. Parallel corpora are an ideal resource for translators, as they contain texts aligned from the source language to the target language \cite{kublerAston2010corporaTranslation,kubler:hal-01134954,bernardini2022corpora}. This makes it relatively easy to perform searches in parallel corpora. However, parallel corpora are complex to compile, as they require bilingual parallel data, which is not easy to obtain for all language pairs and all domains \cite{corpas-pastor-2007-lost,mezeg2020parallel}. Comparable corpora can also be useful for translators \cite{kubler-et-al-corpus-use,KublerMartikainenMestivierPecman+2024+57+78,loock:hal-01391523}. These corpora are independent of each other (not aligned), but contain texts that deal with the same subject area. These corpora are much easier to collect, but searching for bilingual terminology in comparable corpora is more complicated, since data are not aligned. In the educational context of specialised translation training, the use of corpora has also become a central component of the curriculum. It is one of the skills required for translator training within the competence framework of the European Master’s in Translation (EMT) 2022: “Students know how to: make effective use of search engines, corpus-based tools, text analysis tools, computer-assisted translation (CAT) and quality assurance (QA) tools where appropriate.” \cite{emt2022competence}

In this paper, we explore the use of a different class of tools that may partially replace established practices for terminological research in specialised translation. More specifically, we investigate the use of Large Language Models (LLMs) as an alternative to corpus-based approaches traditionally employed to identify translation equivalents. Given that LLMs are trained on massive amounts of textual data and exhibit emerging reasoning and generalisation capabilities, they may provide translators with direct access to plausible terminological translations without the need to manually compile and query corpora. This raises an important question for both professional translators and trainees: to what extent can LLMs reduce—or even eliminate—the need for corpus-based exploration when searching for specialised terminology, and what are the implications of such a shift for translation practices and training?

\section{Related Work}

Since 2022, the year in which ChatGPT, arguably the most popular GenAI model, was released \cite{openai2022chatgpt}, numerous studies have examined the use of LLMs for translation, with very encouraging results \cite{vilar2023promptingpalmtranslationassessing,hendy2023goodgptmodelsmachine,jiao2023chatgptgoodtranslatoryes,wang2023documentlevelmachinetranslationlarge}. Some even believe that the future of machine translation will be closely linked (or is already linked) to the capabilities of LLMs \cite{lyu-etal-2024-paradigm}. Other studies have used LLMs for related tasks, including the evaluation and annotation of translations (both machine-translated and human-translated) \cite{gemba-mqm,gemba,eaprompt,automqm,mt-ranker,instructscore}. With regard to our field of study, only a few studies have explored the potential of LLMs for specialised translation, particularly in the area of terminology.

Recently, \newcite{pecman} looked at the integration of LLMs into terminology analysis, within the framework of the ARTES \cite{artes} term base (a terminology database for research and teaching in specialised translation). This work does not directly address specialised translation, but rather the drafting of definitions for emerging concepts (neologisms) or concepts undergoing semantic change. The study proposes an experimental protocol combining corpus linguistics and interaction with several GenAI tools (ChatGPT, DeepSeek, Perplexity, Claude, Gemini). Pecman \shortcite{pecman} shows that LLMs can be useful for reformulating, improving or structuring terminological definitions, particularly when guided by knowledge-rich contexts extracted from specialised corpora. However, her study emphasises that prior corpus-based analysis remains essential to ensure conceptual accuracy and avoid inaccuracies or approximations. The results highlight that LLMs are a relevant assistance tool, but that they cannot replace a rigorous methodology based on authentic data. 

A few months prior to this study, an experiment on the annotation of specialised translations in an educational context with LLMs, in this case GPT-4o, was conducted \cite{minder-etal-2025-testing}. The researchers prompted the model to annotate errors in students' translations in the Earth sciences domain, based on an error typology adapted to the annotation of specialised translations, the MeLLANGE error typology \cite{castagnoli2011learner}. This typology includes a wide range of terminological errors. After analysing the proportion of errors detected by the LLM, they observed that GPT-4o was able to identify approximately 65\,\% of the terminological errors (specifically 185 out of 289) contained in the Master’s students' translations. This experiment opened up interesting perspectives on the use of LLMs for processing specialised terminology, which led us to conduct the present study.

These findings are particularly relevant to our study, as they already question the role of LLMs in the processing of specialised terminology. However, they focus on derivative tasks (drafting definitions of specialised terms, annotating translation and terminology errors). Our work builds upon this line of thinking by examining more specifically the search for English-French terminology equivalents in specialised translation and by comparing the performance of several LLMs on this task.

\section{Background and Goals}

Our experiment aims to assess the extent to which various GenAI tools can assist specialised translators performing tasks involving terminology research, primarily when searching for equivalents from English to French. This question is particularly relevant in specialised translation, where terminology accuracy plays a critical role in the quality, reliability and domain appropriateness of the target text.

While LLMs are increasingly explored and used for translation and MT evaluation tasks, their actual usefulness for more specific terminological tasks remains, to our knowledge, poorly documented, especially in LSP with a focus on specialised translation. Unlike the vast majority of studies on LLMs, which tend to focus on general language, our research is grounded in two highly specialised and contrasting fields: Earth, Environmental and Planetary Sciences (EEPS) and Natural Language Processing (NLP). These two domains were intentionally selected to test the performance of LLMs across diverse and varied knowledge areas. By comparing the results across these two domains, we hope to determine whether the models tested here remain stable or vary depending on the domain. Beyond a simple comparison of models' performances, the ultimate goal is more practice-oriented. We aim to explore whether LLMs can realistically serve as assistants for specialised translators in terminology research, or even whether LLMs have the potential to replace corpora.

\section{Methods}

\subsection{Models}
In order to assess the potential usefulness of LLMs for specialised terminology research, we examine the performance of four models in the search for equivalents from English into French: GPT-4o, GPT-5.2, Claude Sonnet 4.5 and DeepSeek. We selected these models for pragmatic reasons. As this experiment aims to assess the potential usefulness of LLMs for finding terminological equivalents, we intend to test the tools that are exploited by users. These proprietary models are used in practice by professional translators and translation learners. We test the models on 80 terms per domain, for a total of 160 terms per model. 

\begin{figure*}
\centering
\includegraphics[width=.9\textwidth]{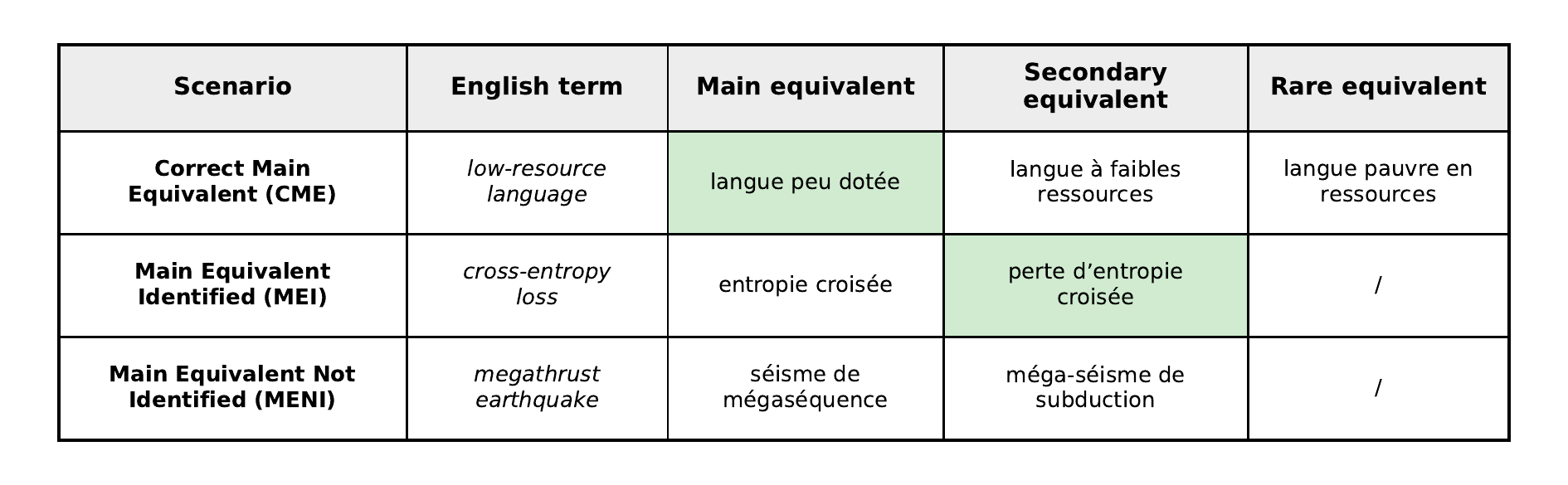}
\caption{Suggestions from LLM – examples for each case: main equivalent correctly identified as main equivalent by the LLM (first example); main equivalent identified as a secondary equivalent by the LLM (second example); main equivalent not identified by the LLM (third example). A green cell indicates that the equivalent is considered appropriate by the professional translator. In the third example, the model failed to identify the correct French equivalent; the French term is \textit{séisme de mégachevauchement} (ou \textit{séisme de méga-chevauchement}). \label{fig:scenarios}}
\end{figure*}

\subsection{Term selection}
The first step was to select 80 terms per domain (EEPS and NLP), for a total of 160 terms. To do so, we manually selected terms from several specialised resources, namely texts (research articles) translated by Master's students in specialised translation, a terminology database, ARTES\footnote{https://artes.app.univ-paris-diderot.fr/}, that is fed annually by, among others, Master's students, and other specialised texts that we had used for other experiments on specialised translation (references will be provided once the submission can be de-anonymised). We ensured that terms of different structures and varying levels of complexity were represented in our sample: simple terms as well as compound and complex terms. The 80 terms are identical for all models and prompts. The selected terms are detailed in Figures~\ref{fig:terms-nlp} and~\ref{fig:terms-eeps} in the appendices; the responses generated by each LLM for each domain and mode, along with the annotations for these responses, will be made available upon publication. 

\subsection{Prompts and modes}
We test two different prompts for each model and each domain. These prompts correspond to different modes. The first prompt is what we call ‘terminology mode’: it simply instructs the model to search for terminological equivalents, and for each term, we provide in the prompt a context sentence (in English) containing the term. In the second mode, ‘translation mode’, we first request the model to translate the context sentence provided, then to list the identified terminology equivalents (Figures~\ref{fig:prompt-term} and~\ref{fig:prompt-trans} in the appendices illustrate the specific differences between terminology mode and translation mode). A total of 16 individual experiments were conducted (4 models, 2 domains and 2 modes). In both prompts, we also request LLMs to indicate their level of confidence for each term processed. In addition, we request LLMs to sort the equivalents found into three categories: main equivalent, and, where applicable, secondary equivalents and rare equivalents. The criteria we give to the models are: “(a) main equivalent – attested (equivalent found in one or more terminology databases) and/or more frequent (frequently found in similar contexts in corpora, in texts of the same register and type); (b) secondary equivalents – less frequent in terminology databases and corpora, but still used; (c) rare equivalents (found only occasionally and with statistically insignificant occurrences). If there is only one equivalent, list it as the main equivalent. There do not necessarily have to be multiple equivalents” (excerpt from the prompt). Two examples have also been added to the prompt to show the desired output presentation.

In addition, we introduced a last variant (referred to as ‘documentary justification’ in the following) on the best-performing model. On this model, we tested the inclusion of a constraint: we asked the model, for each equivalent it identified, to provide a source (scientific, which could be an article, a book chapter, a conference paper, etc.) containing that equivalent and to give a sentence illustrating the use of the equivalent. We explicitly instructed the model not to invent any sources or sentences. Our idea was to see whether forcing the model to identify a source would yield more accurate and precise results. The full prompts for both modes and for the documentary justification are detailed in Figures~\ref{fig:prompt-term} and~\ref{fig:prompt-trans} in the appendices.

\subsection{Performance, annotation and evaluation}
For each term, one professional translator with extensive expertise in corpus linguistics and translation annotation in both domains  analyses whether the LLM is able to provide a correct French equivalent. This allows us to precisely quantify the proportion of equivalents correctly identified by each model, and to compare the impact of the prompting strategy and the domain of specialisation.

For performance evaluation, we first classified each equivalent provided by the model as follows. Each term was assigned one of the possible outcomes: the main equivalent was correctly identified as the main equivalent, the main equivalent was identified but classified as a secondary or rare equivalent, or the main equivalent was not identified (see Figure~\ref{fig:scenarios} for an example of each case). These three outcomes were weighted differently in the scoring procedure: full credit was assigned when the main equivalent was correctly identified as the main equivalent, partial credit when it was identified but not ranked as the main equivalent, and no credit when it was not identified. The final score for each model, mode and domain was then obtained by averaging these weighted outcomes across all terms and normalising the result on a 0 to 1 scale.

\subsection{Verification resources}
In order to check LLM answers, we use several resources: comparable and parallel corpora (English-French) compiled and enriched over the years in our research lab as part of various projects, specialised terminology databases such as TERMIUM\footnote{https://www.btb.termiumplus.gc.ca/.}, and a terminology database developed in our research lab. (We will provide proper references for these resources when the authors' identities may be disclosed). All answers, for the 80 terms per domain, each model and each prompting method, were annotated manually.

\section{Results}

We analysed the models' performance across different modes and domains according to several statistical indicators.
\subsection{Overall performance}

\begin{figure}[htbp]
    \centering
    \includegraphics[width=\columnwidth]{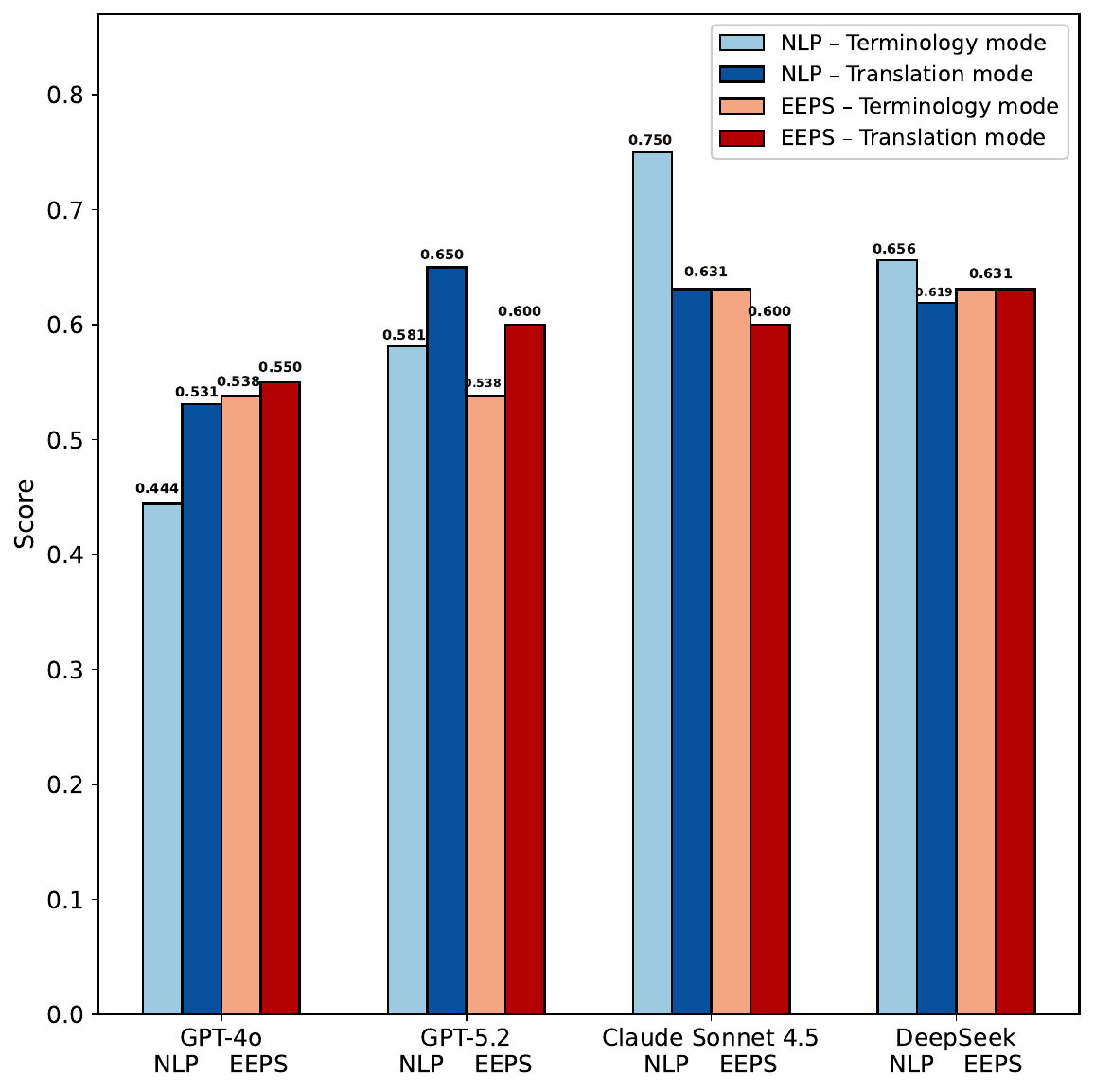}
    \caption{Results achieved on the four models, by mode (terminology or translation) and by domain (EEPS and NLP).}
    \label{fig:overall-results}
\end{figure}

\begin{figure*}
\centering
\includegraphics[width=.9\textwidth]{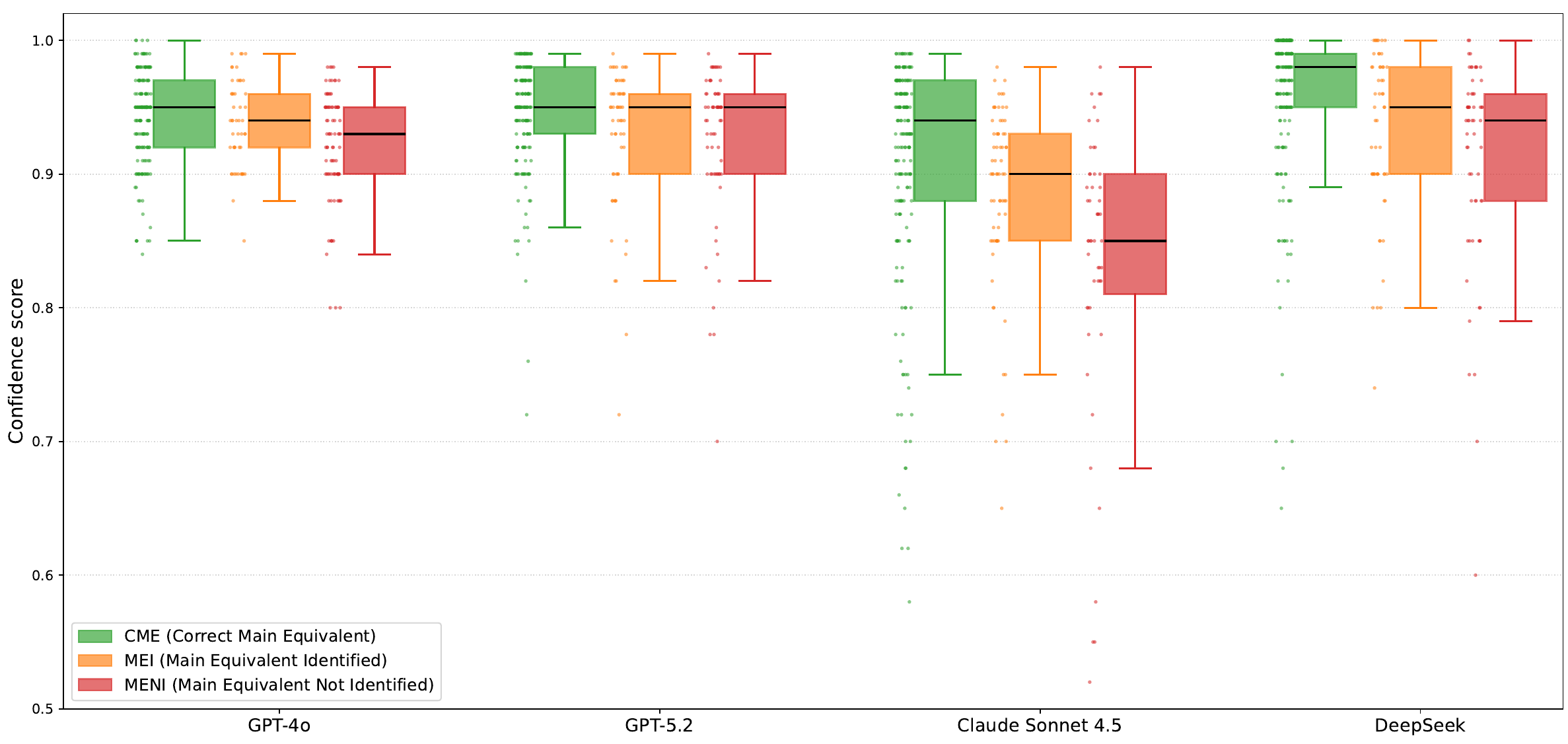}
\caption{Distribution of confidence estimates for each model, based on the accuracy of the identified equivalents (green: the main equivalent has been identified as the main equivalent; orange: the main equivalent has been identified as a secondary or rare equivalent; red: the main equivalent has not been identified). \label{fig:distrib-conf-accuracy}}
\end{figure*}

Figure~\ref{fig:overall-results} shows that, overall, there are substantial variations in results across models, modes and, to a lesser extent, domains. Of the four LLMs tested, Claude Sonnet 4.5 achieves the highest average score, with peak performance in terminology mode in the NLP domain. However, Claude Sonnet 4.5 shows greater variation across domains and modes than the other models. DeepSeek performs satisfactorily and shows stable results across modes and domains. GPT-5.2 achieves intermediate scores, and GPT-4o remains the weakest model overall for this experiment.

The effect of the prompting strategy (terminology vs translation mode) is not consistent across models. GPT models seem to benefit from translation mode, with a systematic improvement from terminology to translation mode. On the other hand, Claude Sonnet 4.5 performs better in terminology mode, particularly in NLP. DeepSeek shows only very limited variation depending on the mode, suggesting potential robustness to the prompting strategy.

The effects of domain are also noticeable, but less consistent than the effects of mode and prompt. Some models perform slightly better in NLP (GPT-5.2 and Claude Sonnet 4.5), while GPT-4o performs slightly better in STEP. DeepSeek, on the other hand, remains fairly stable across both domains. This suggests that the domain does influence model performance, but that this effect interacts with other conditions: the model and the prompting strategy. DeepSeek, while not achieving the best scores, is the most stable model to changes in mode and domain. Although the difference in performance across different domains appears to be rather minimal, this may be useful for specialised translators: selecting a specific model based on the domain covered may yield better results.

These results do not allow for the conclusion that LLMs can replace corpus-based searches by specialised translators. However, the results are sufficiently positive to suggest that LLMs can indeed provide useful support in the search for equivalents, with some models performing slightly better in one area than in another, in translation mode or terminology mode. 

\subsection{Confidence estimates}

Figure~\ref{fig:distrib-conf-accuracy} shows the distribution of confidence estimates reported by each model, compared in terms of the accuracy of equivalents (identified (green), partially identified (yellow) and unidentified (red)). Overall, confidence estimates remain high across all three categories. However, there is a general trend across all four models towards higher average confidence estimates when the main equivalent is correctly identified. In contrast, partially identified and unidentified equivalents tend to be associated with lower average confidence estimates. This trend is particularly noticeable in Claude Sonnet 4.5, where the distribution of estimates becomes progressively weaker and more dispersed as accuracy drops. In other words, when Claude Sonnet 4.5 fails to correctly identify the main equivalent, it reports a generally lower confidence estimate than the other models. GPT-4o, GPT-5.2 and, to a lesser extent, DeepSeek, however, tend to remain confident in all three scenarios. Consequently, for these models, confidence estimates are less clearly aligned with terminological accuracy. From a practical perspective, these results show that confidence estimates are only partially informative as indicators of terminological accuracy.

\begin{figure*}
\centering
\includegraphics[width=.9\textwidth]{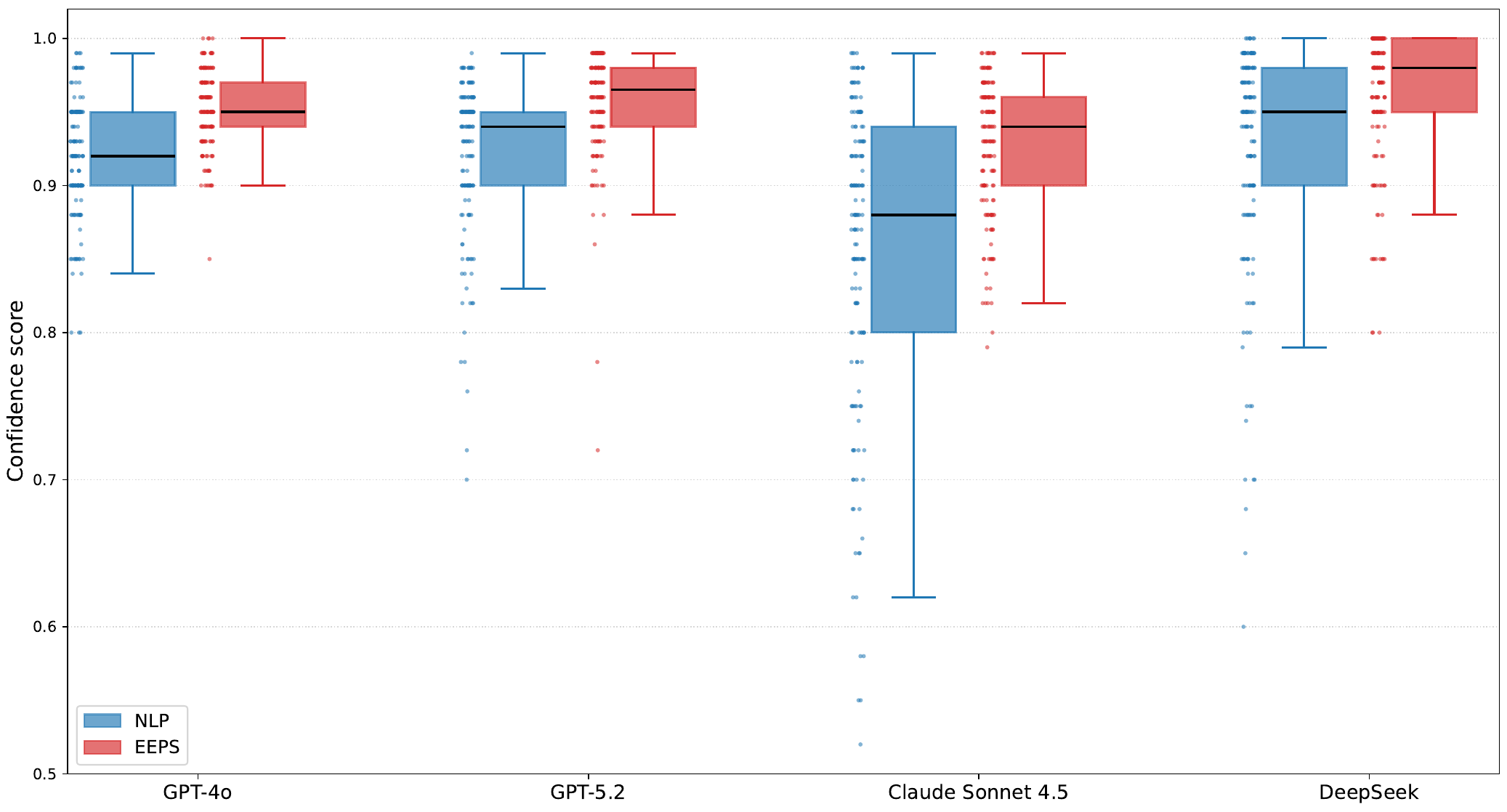}
\caption{Distribution of confidence estimates for each model, based on the domains (NLP in blue vs EEPS in red). \label{fig:distrib-conf-dom}}
\end{figure*}

Figure~\ref{fig:distrib-conf-dom} shows that all four LLMs report higher confidence levels in the EEPS domain, although confidence scores also remain relatively high in NLP. Claude Sonnet 4.5 has a more distinct profile: it has a significantly more dispersed and weaker distribution in NLP, showing that the NLP domain is associated with the presence of many cases of lower confidence. However, Figure 2 shows that Claude Sonnet 4.5 is one of the models with the best results in NLP, particularly in terminology mode, further demonstrating that confidence estimates do not necessarily correlate with accuracy.

From the user’s perspective, these results show that confidence estimates alone do not constitute a reliable indicator of terminological accuracy. While there is a slight overall trend--correctly identified equivalents are, on average, accompanied by slightly higher confidence estimates--the significant overlap between the three categories shows that high estimates can also be associated with partially identified, or even unidentified, equivalents. In other words, a model may be highly confident while proposing an inadequate equivalent. From a practical perspective, this means that translators cannot rely on the confidence estimate alone to judge the probability that a suggested equivalent is correct. At best, this estimate can serve as a secondary indicator, but it cannot replace verification in corpora, term bases or other resources.

\subsection{Documentary justification}

Based on the previous results, we considered Claude Sonnet 4.5 to be the best-performing model according to several variables: good overall performance in both domains (Figure 2) and a stronger correlation between confidence estimates and accuracy of equivalents (Figure 3). Consequently, we tested the addition of an instruction in the prompt for this model: providing a scientific source and a sentence containing the term from that source. The purpose was to determine whether forcing the model to rely on concrete references would yield better results. We only tested this in terminology mode, as this is the mode in which Claude Sonnet 4.5 achieves the best results.

\begin{figure}[htbp]
    \centering
    \includegraphics[width=\columnwidth]{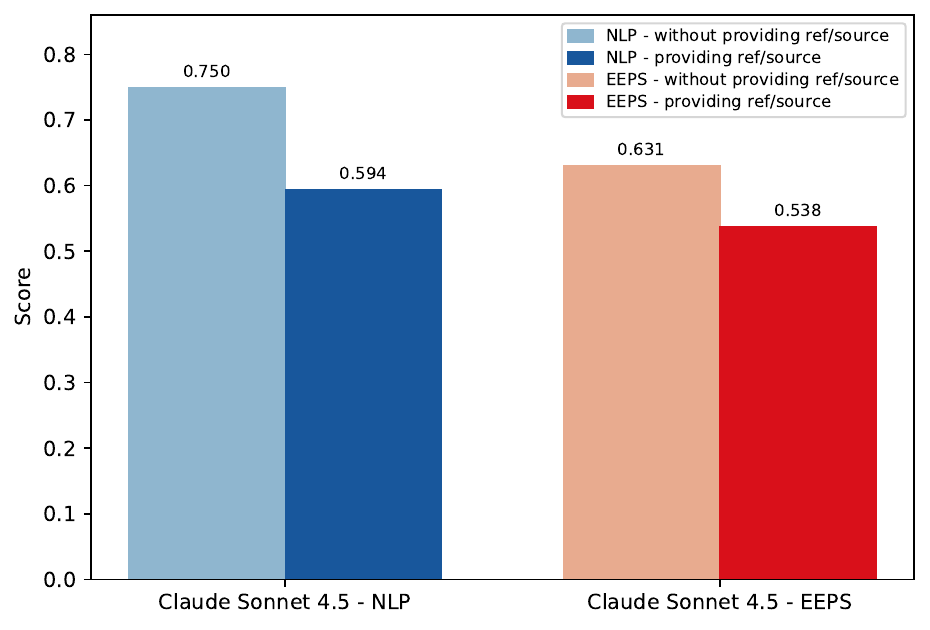}
    \caption{Comparison of the performance of Claude Sonnet 4.5 in terminology mode, with and without the specification of a scientific reference and a sentence illustrating the term.}
    \label{fig:claude-ref}
\end{figure}

Figure~\ref{fig:claude-ref} shows the results obtained with Claude Sonnet 4.5 in terminology mode, with and without the addition of an instruction requiring the model to provide, for each equivalent identified, a scientific source and an example sentence from that source. In both domains, the addition of this constraint leads to a decrease in scores. The decrease is therefore visible in both domains, and particularly marked in NLP. This approach to documentary justification poses an additional issue. Although we explicitly instructed the model not to invent any references or sentences illustrating the term, this instruction was not observed. In most cases, the sources provided by the model do not exist: they are either completely fabricated or are a translated source (for example, the source does not exist in French as provided by Claude Sonnet 4.5, but can be found in English). Furthermore, in every single case without exception, the sentence given as a reference by the model is made up, meaning that, upon verification, we cannot find it anywhere.
These results show that, in the framework of this experiment, adding a requirement to reference and justify with a source does not improve the model's performance in finding equivalents. On the contrary, this additional constraint seems to be associated with a decrease in the model's ability to correctly identify the main equivalent. However, it is important to provide some insight into this observation. This drop in performance may not necessarily be (solely) due to the addition of the new instruction. Indeed, LLMs' performances are unpredictable: it has been observed that the performance of models varies between different iterations of the same model \cite{siu2023chatgptgpt4translators}. It is therefore not directly possible to claim that this decline in performance is the result of the addition of the documentary justification instruction.

\subsection{Qualitative analysis}
Beyond quantitative and statistical analysis, it is also interesting to observe in concrete terms how the equivalents provided by LLMs, even when incorrect, can potentially guide a translator towards the right solution. Other examples show, on the contrary, that there are cases—albeit rare—where LLMs produce completely nonsensical or even absurd results.

\begin{figure}[htbp]
    \centering
    \includegraphics[width=\columnwidth]{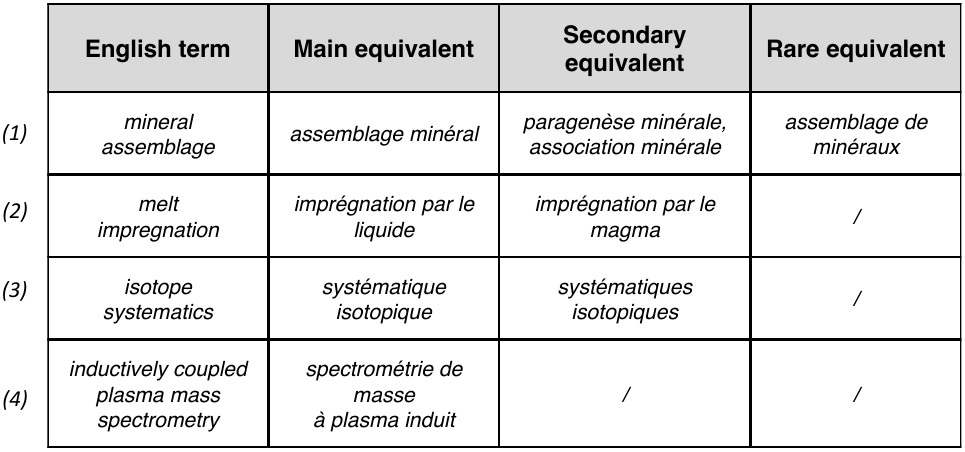}
    \caption{Examples of equivalent search results by LLMs that are inaccurate but may still prove useful to a specialised translators. The slash (/) in ‘Secondary equivalent’ and ‘Rare equivalent’ indicates that the LLM did not identify any equivalents of this type, and that it identified only a main equivalent, with a secondary equivalent where applicable.}
    \label{fig:examples}
\end{figure}

The examples illustrated in the Figure~\ref{fig:examples} show instances where the equivalents provided by LLMs are inaccurate, but where a simple corpus or term base search based on the equivalents suggested by the LLM quickly provides an accurate solution. The first example (\textit{mineral assemblage}) is the most straightforward: a simple search for \textit{assemblage} associated with the stem \textit{minéral} shows that the most common term is \textit{assemblage minéralogique}, i.e. the same head associated with an adjective that has the same stem as \textit{minéral}. Here, the LLM does not provide the correct equivalent, but it does provide clear clues pointing to the correct solution.

Lines 2 and 3 are similar in terms of approach: they are both examples of confusion between attributive adjectives and noun complements (compléments du nom) in French. In French, modifiers can take several forms: either an attributive adjective (an adjective directly linked to the noun, without a preposition), or a noun complement (a modifier separated from the noun by a preposition). Here, \textit{imprégnation par le magma} and \textit{systématique isotopique} fall into this category: a quick corpus query for the head nouns (\textit{imprégnation} and \textit{systématique}) combined with the stems \textit{magma} and \textit{isotop*} provides the answer. This reveals the most frequently used terms: \textit{imprégnation magmatique} and \textit{systématique des isotopes}.

Example 4 is even more straightforward. It is sufficient to search for \textit{spectrométrie de masse} (confirmed in term bases, including Termium) in association with \textit{plasma}, and this should lead to the solution: \textit{spectrométrie de masse à plasma	à couplage inductif}.

These examples demonstrate that even when the LLM does not provide the appropriate equivalents, the suggestions can guide a translator in conducting the correct searches (in corpora or term bases). However, this depends on several factors, including experience in the domain (particularly the intuition that can be acquired with experience) and with corpus exploitation tools. Therefore, it should not be assumed that these examples would be useful for any given translator. Only practical experiments in real-world contexts would allow us to assess the usefulness of LLM suggestions.

\begin{figure}[htbp]
    \centering
    \includegraphics[width=\columnwidth]{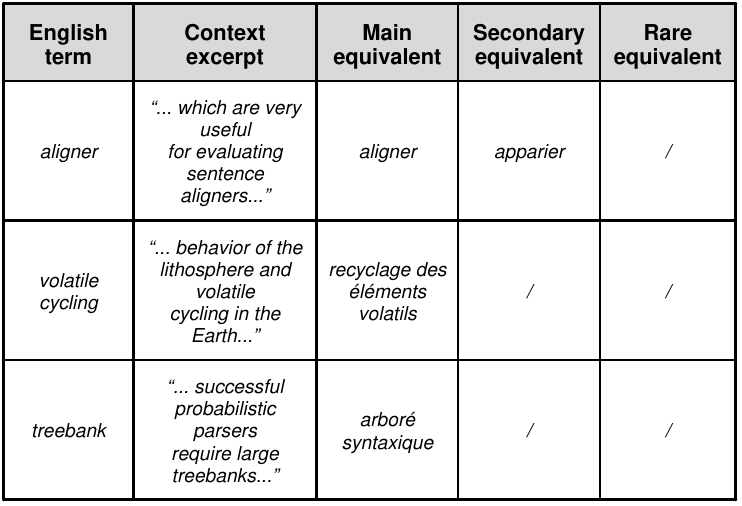}
    \caption{Examples of LLM hallucinations when searching for equivalents. The slash (/) in ‘Secondary equivalent’ and ‘Rare equivalent’ indicates that the LLM did not identify any equivalents of this type, and that it identified only a main equivalent, with a secondary equivalent where applicable.}
    \label{fig:hallucinations}
\end{figure}

Figure~\ref{fig:hallucinations} shows striking examples of hallucinations produced by LLMs in the search for equivalents. The first term, \textit{aligner}, which is a noun referring to an alignment tool, is associated with a verbal equivalent, i.e. the verbs \textit{aligner} (to align) and \textit{apparier} (to pair up) in French;  the issue here is that the LLM did not accurately identify the part-of-speech of the term, despite the context sentence clearly illustrating the noun. The correct equivalent here would be \textit{aligneur} or \textit{outil d'alignement} In the second example, the term \textit{cycling} was translated as \textit{recyclage} (recycling) in French, which does not correspond at all to the meaning of the English term. A correct French equivalent for this term would be \textit{cycle des (éléments) volatils} ou \textit{cycle des espèces volatiles}. In the last example, the noun has been translated as an adjectival phrase without a head noun, which makes no sense. The correct equivalent for this term is \textit{corpus arboré}. These three examples show that, although in most cases LLMs identify the correct equivalents or provide good starting points for the search of the correct equivalent, they can still produce completely disconnected or absurd outputs.

\section{Discussion}

The results show that LLMs can prove useful tools for specialised terminology research, but that, within the current scope of this experiment, they cannot replace specialised corpora. Even in the best configurations, performance remains limited and uneven depending on the model, mode and domain. These results therefore argue in favour of using LLMs as a complementary tool in specialised translation workflows rather than as a direct substitute for comparable, parallel corpora or terminology databases. LLMs can help to quickly generate potential equivalents, but the verification, validation and contextualisation of terms remain the responsibility of external resources and human expertise.

The effect of the model appears to be the most decisive factor. Claude Sonnet 4.5 achieves the best overall results in the most favourable configuration (terminology mode), while DeepSeek stands out for its greater stability between modes and domains. GPT-5.2 occupies an intermediate position and GPT-4o lags behind overall. In addition, the effect of the prompting strategy is not consistent: GPT models seem to benefit from the translation mode, while Claude Sonnet 4.5 performs better in terminology mode. This suggests that there is no universally ideal prompting strategy for terminological equivalence search, and that effectiveness largely depends on the model being queried.

Analysis of confidence estimates also shows that these are only a partial indicator of terminological accuracy. While Claude Sonnet 4.5 seems to adjust its confidence better when the actual quality of the equivalent decreases, LLMs often maintain high confidence levels, even when the main equivalent is partially identified or unidentified. This highlights possible overconfidence and suggests that self-reported confidence should not be considered a reliable indicator of terminological accuracy. The domain effect also comes into play, but remains moderate and inconsistent. Confidence levels appear to be slightly higher and more stable in EEPS than in NLP for several models, although this trend cannot be generalised to all cases.

Finally, the complementary experiment conducted on Claude Sonnet 4.5 shows that adding a requirement to provide a scientific source and an example sentence does not improve performance, but rather degrades it in both NLP and EEPS. This finding suggests that an additional constraint of documentary justification does not guarantee better terminological quality and may even distract the model from the main task.

\section{Conclusion and Future Work}

Specialised translation relies heavily on the use of documentation and terminology resources, among which corpora play a central role. However, compiling and exploiting them has significant drawbacks: it requires time, specific technical skills and access to (parallel) data that can sometimes be difficult to obtain, particularly for certain domains and language pairs. It is in this context that this study sought to assess the extent to which proprietary LLMs available to general, non-technical users can assist specialised terminology research.

Based on four models tested on a task of finding English-French equivalents in two specialised domains, the results show that LLMs have real potential for assistance, but that their performance varies depending on the model, the prompting strategy and, to a lesser extent, the domain. They also show that the confidence estimates provided by the models are only a partial indicator of terminological accuracy. Overall, our results therefore support the idea of LLMs playing a complementary role in specialised translation workflows rather than directly replacing specialised corpora.

This study does, however, have several limitations. It focuses on two specialised fields and a single language pair, English-French, and on a specific experimental task of finding equivalents. In future work, it could be useful to extend the analysis to other fields, other languages and other prompting configurations. Another limitation of this study is that it is based on annotations by a single expert, with no measurement of inter-annotator agreement between different expert annotators. This is due to the highly specialised, complex and time-consuming nature of this task. In the future, we aim to involve several expert annotators and incorporate IAA measurements, either across the entire dataset or a subset of the data. This study also lacks testing to determine whether using LLMs as a support tool can reduce time and cognitive load for specialised translators, as opposed to relying, as usual, on corpora and terminology databases. The qualitative analysis of LLM outputs has been undertaken in this study, but in future work, we intend to assess in real-world settings (such as educational contexts involving translation learners) how these outputs can be exploited and how they might complement or be combined with corpus-based alternatives.
Most importantly, it would be relevant to go beyond task evaluation to examine more practically the usefulness and usability of these tools for specialised translators, particularly in an educational context, for example with Master's students in specialised translation. Such investigations would make it possible to observe more concretely their effect on the time spent searching for terminology, the quality of the choices made, the verification strategies implemented and, more broadly, their place in translation and training practices.

\section*{Acknowledgments}
This research was funded by the French Agence Nationale de la Recherche (ANR) under the project MaTOS - ``ANR-22-CE23-0033-03''.

\bibliographystyle{eamt26}
\bibliography{eamt26}
\clearpage
\onecolumn

\section{Appendices}
\subsection{Terms selected (NLP)}

\begin{figure}[!ht]
\centering
\includegraphics[width=.9\textwidth]{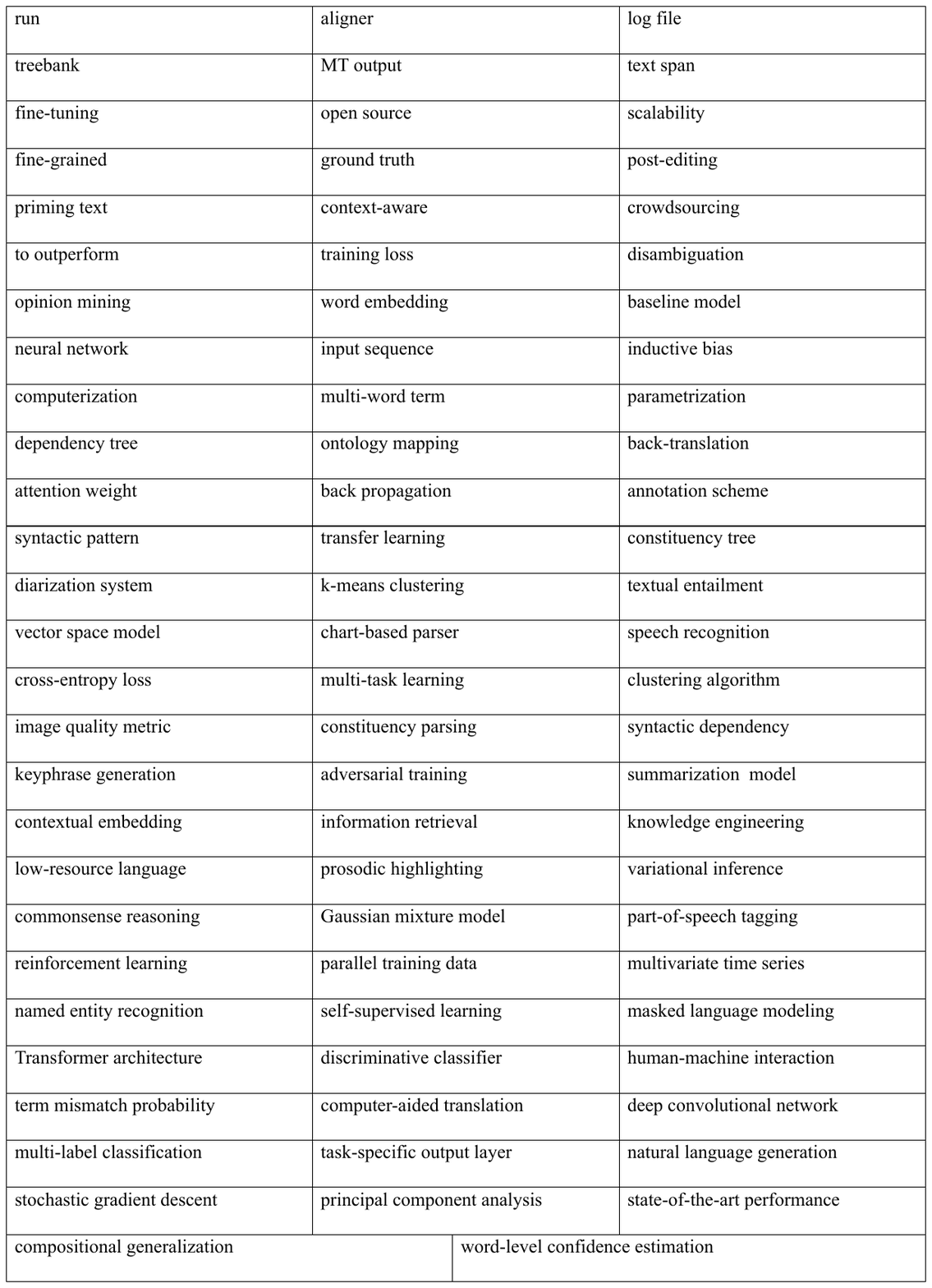}
\caption{80 terms selected for the Natural Language Processing domain.\label{fig:terms-nlp}}
\end{figure}

\clearpage
\subsection{Terms selected (EEPS)}

\begin{figure}[!ht]
\centering
\includegraphics[width=.9\textwidth]{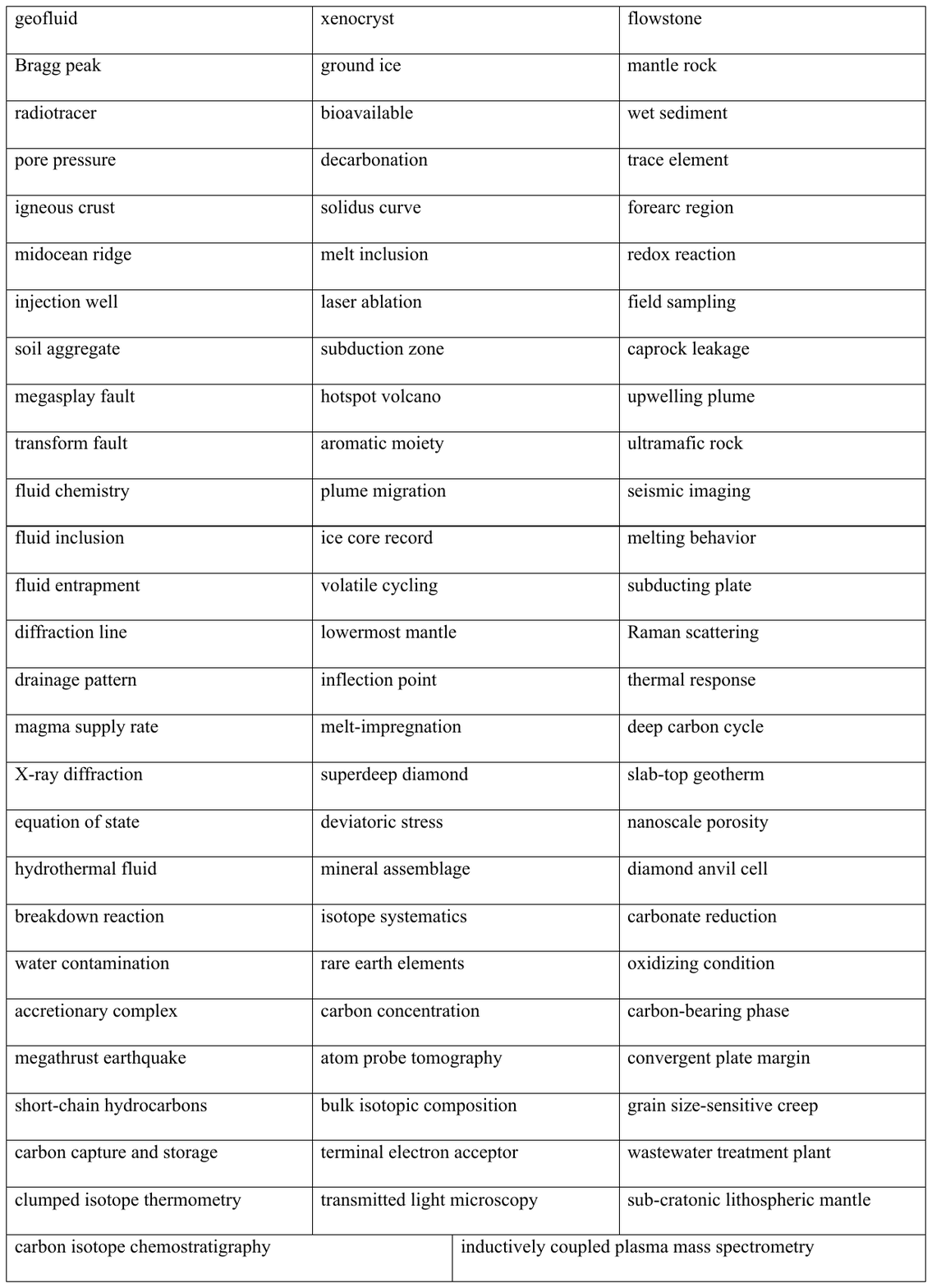}
\caption{80 terms selected for the Earth, Environmental and Planetary Sciences domain.\label{fig:terms-eeps}}
\end{figure}

\clearpage
\subsection{Prompt: terminology mode}
\begin{figure}[!ht]
\centering
\includegraphics[width=.9\textwidth]{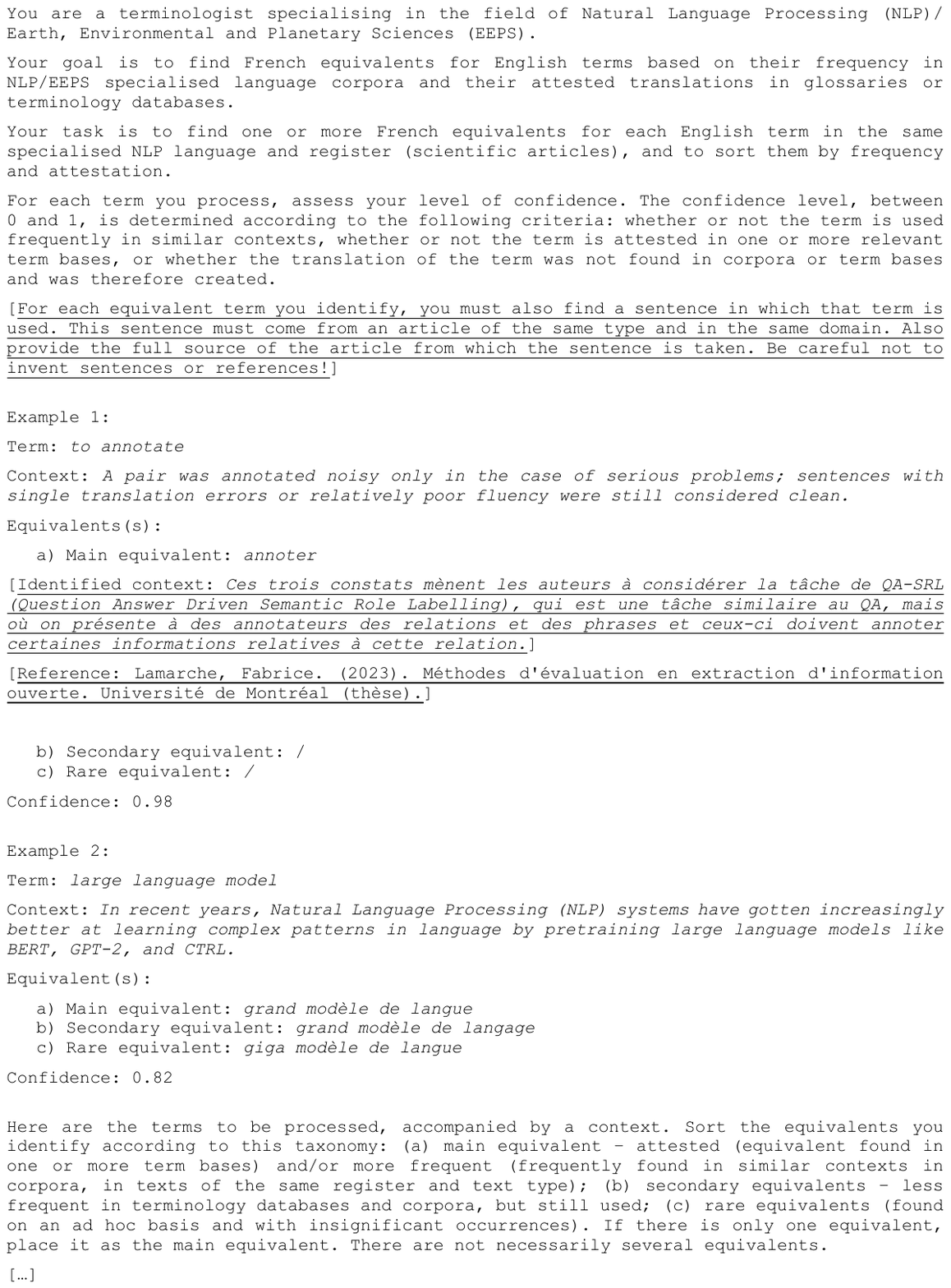}
\caption{Prompt used for terminology research on LLMs in “terminology” mode. The segments in square brackets and underlined were added only for the additional documentary justification test on Claude Sonnet 4.5.\label{fig:prompt-term}}
\end{figure}

\clearpage
\subsection{Prompt: translation mode}
\begin{figure}[!ht]
\centering
\includegraphics[width=.9\textwidth]{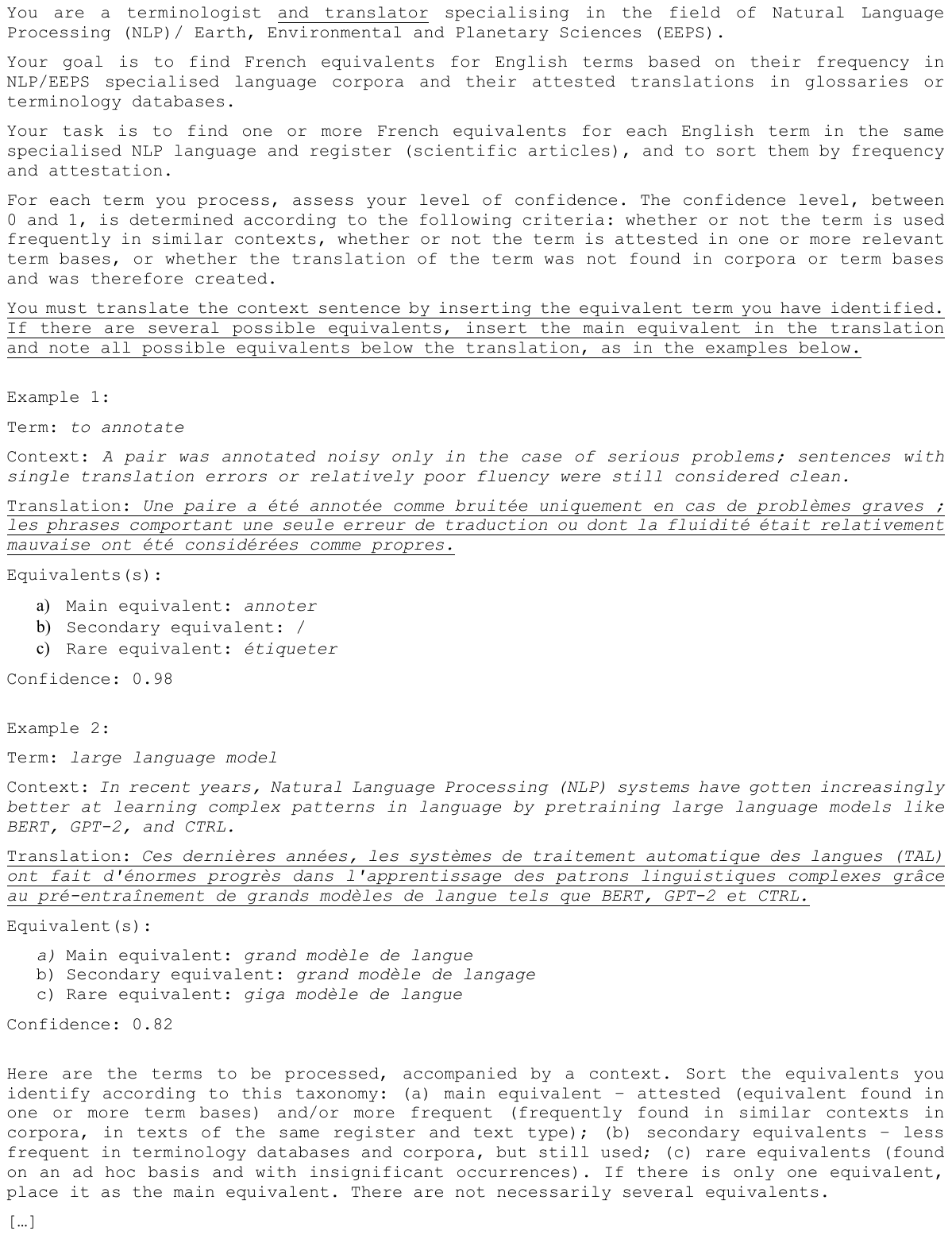}
\caption{Prompt used for terminology research on LLMs in “translation” mode. The underlined segments are those that have been added for ‘translation’ mode compared to ‘terminology’ mode.\label{fig:prompt-trans}}
\end{figure}

\end{document}